%% file: article.tex
\documentclass[sn-mathphys,Numbered]{sn-jnl}% Math and Physical Sciences 

\usepackage{graphicx}%
\usepackage{multirow}%
\usepackage{amsmath,amssymb,amsfonts}%
\usepackage{amsthm}%
\usepackage{mathrsfs}%
\usepackage[title]{appendix}%
\usepackage{xcolor}%
\usepackage{textcomp}%
\usepackage{manyfoot}%
\usepackage{booktabs}%
\usepackage{algorithm}%
\usepackage{algorithmicx}%
\usepackage{algpseudocode}%
\usepackage{listings}%

\usepackage{lmodern}
\usepackage{float}
\usepackage{subfig}
\DeclareMathAlphabet{\mathbbold}{U}{bbold}{m}{n}

\begin{document}

\title[Article Title]{Sashimi-Bot: Autonomous Tri-manual Advanced Manipulation and Cutting of Deformable Objects}


\author*[2]{\fnm{Sverre} \sur{Herland}}\email{sverre.herland@sintef.no; ekrem.misimi@sintef.no}
\equalcont{These authors contributed equally to this work.}

\author[1]{\fnm{Amit} \sur{Parag}}%\email{iiauthor@gmail.com}
\equalcont{These authors contributed equally to this work.}

\author[1]{\fnm{Elling Ruud} \sur{Øye}}%\email{iiauthor@gmail.com}
\equalcont{These authors contributed equally to this work.}

\author[3]{\fnm{Fangyi} \sur{Zhang}}%\email{iiauthor@gmail.com}
\equalcont{These authors contributed equally to this work.}

\author[4]{\fnm{Fouad} \sur{Makiyeh}}%\email{iiauthor@gmail.com}
\equalcont{These authors contributed equally to this work.}

\author[1]{\fnm{Aleksander} \sur{Lillienskiold}}
\equalcont{These authors contributed equally to this work.}

\author[6]{\fnm{Abhaya} \sur{Pal Singh}}

\author[5]{\fnm{Edward H.} \sur{Adelson}}

%
%\author[1,2]{\fnm{Second} \sur{Author}}\email{iiiauthor@gmail.com}

\author[4]{\fnm{François} \sur{Chaumette}}%\email{iiiauthor@gmail.com}

\author[4]{\fnm{Alexandre} \sur{Krupa}}%\email{iiiauthor@gmail.com}

\author[3]{\fnm{Peter} \sur{Corke}}%\email{iiiauthor@gmail.com}

\author*[1]{\fnm{Ekrem} \sur{Misimi}}%\email{iiiauthor@gmail.com}

\affil*[1]{\small{\orgdiv{Robotics and Intelligent Automation}, \orgname{SINTEF Ocean}, \orgaddress{\city{Trondheim}, 
\country{Norway}}}}

\affil*[2]{\small{\orgdiv{Department of Computer Science}, \orgname{NTNU - Norwegian University of Science and Technology}, \orgaddress{\city{Trondheim},
\country{Norway}}}}

\affil[3]{\small{\orgdiv{QUT Centre for Robotics}, \orgname{Queensland University of Technology}, \orgaddress{\city{Brisbane}, 
\country{Australia}}}}

\affil[4]{\small{\orgdiv{Rainbow}, \orgname{Inria, University of Rennes, CNRS, IRISA}, \orgaddress{\city{Rennes},  \country{France}}}}

\affil[5]{\small{\orgdiv{CSAIL}, \orgname{MIT - Massachusets Institute of Technology}, \orgaddress{\city{Boston}, \country{USA}}}}

\affil[6]{\small{\orgdiv{RealTek}, \orgname{Norwegian University of Life Sciences (NMBU)}, \orgaddress{\city{Ås},  \country{Norway}}}}

\abstract{Advanced robotic manipulation of deformable, volumetric objects remains one of the greatest challenges due to their pliancy, frailness, variability, and uncertainties during interaction. 
Motivated by these challenges, this article introduces Sashimi-Bot, an autonomous multi-robotic system for advanced manipulation and cutting, specifically the preparation of sashimi. 
The objects that we manipulate, salmon loins, are natural in origin and vary in size and shape, they are limp and deformable with poorly characterized elastoplastic parameters, while also being slippery and hard to hold. 
The three robots straighten the loin; grasp and hold the knife; cut with the knife in a slicing motion while cooperatively stabilizing the loin during cutting; and pick up the thin slices from the cutting board or knife blade. 
Our system combines deep reinforcement learning with in-hand tool shape manipulation, in-hand tool cutting, and feedback of visual and tactile information to achieve robustness to the variabilities inherent in this task. 
This work represents a milestone in robotic manipulation of deformable, volumetric objects that may inspire and enable a wide range of other real-world applications.
}


\maketitle

\input{sec/1_introduction}
\input{sec/2_system}

\input{sec/3_results}
\input{sec/4_discussion}

\input{sec/5_content}

\bibliography{references}% common bib file

\begin{appendices}

\input{app/1_methods}


\end{appendices}


\end{document}

%% file: sec/1_introduction.tex
\section{Introduction}
\label{sec:introduction}

\newcommand\ekremm[1]{{\textcolor{blue}{#1}}} %
\newcommand\krupa[1]{{\textcolor{cyan}{#1}}} %
\newcommand{\AS}[1]{\textcolor{brown}{#1}}

This article presents Sashimi-Bot, a multi-robot system that can autonomously prepare sashimi.
Its three robotic arms collaborate to transform a salmon loin, placed arbitrarily on a cutting board, into a set of slices arranged neatly on a serving tray.  

We chose this task as an exemplar of a useful but complex realworld task that involves the manipulation of deformable objects. Robot manipulation is often considered to be simply the grasping of rigid objects. In constrast, Sashimi-bot performs three very different types of manipulation of non-rigid (elastoplastically deformable) material: straigthening, cutting and picking up. The objects that we manipulate, salmon loins, are natural in origin, vary in size and shape, are limp and deformable with poorly characterized elastoplastic parameters, and are slippery and hard to hold.  Research into robotic manipulation of such objects is scant in the literature.

Adding to the challenge, the operations must be achieved reliabily despite the inherent uncertainty, and the object must be treated gently so as to not damage it.  
The quest to enhance robots with advanced manipulation capabilities rivaling those of humans is a long-running endeavor.
The era of Deep Learning has been an essential driver for many recent advances, although with limitations \cite{sünderhauf2018limits}, and addressing the intricacies of manipulating deformable objects remains a persistent challenge~\cite{ma2023, zhu2022challenges,misimi2018robotic, lee2021learning}. 
While significant progress has been achieved in diverse domains such as tactile-base learning \cite{zhao2023}, shape-servoing \cite{makiyeh2023}, advanced capabilities in manipulation \cite{thach2022learning,shi2022robocraft,sivertsvik2024}, and cutting \cite{xu2023}, endowing robots with a plethora of new skills and greater autonomy \cite{fu2023mobilealoha} is an ongoing pursuit.

Humans, however, routinely manipulate deformable objects, and with a little practice, can become adept at sashimi slicing.
With one hand, we manipulate the loin into a position and shape suitable for cutting while grasping a knife with the other. 
We then carefully cut the loin into thin slices using a knife held in one hand while stabilizing the loin with our free hand which applies a gentle grasp.
Finally, we pick up the slices using our fingers or some appropriate utensils and arrange them on a serving tray.  We adapt effortlessly to variations in loin thickness and texture.
Our system mimics this human-like approach.  The three robots perform the shape manipulation, knife holding and cutting, gentle stabilizing grasp, and slice pickup.  They use a number of feedback loops, driven by visual and tactile information, to ensure that each manipulation stage is driven to completion despite uncertainty.  Visual feedback of manipulation is the robotic equivalent of human hand-eye coordination, while the tactile feedback is the robotic equivalent of human touch perception.

On a grand scale, robotic manipulation holds the potential to unlock transformative changes across key sectors, driving advancements in sustainable development \cite{guenat2022sustainable}, food production \cite{lipson2023robots} and security \cite{hodson2017food}, and resource efficiency \cite{misimi2016gribbot}.
It is widely accepted that manipulation remains one of the greatest challenges in robotics \cite{Cui2021NextGenManipulation}, due to the need to manage physical contact with objects and the environment, as well as the interaction mechanisms involved \cite{Cui2021NextGenManipulation}. 
Deformable object manipulation stands out as an especially difficult problem to solve \cite{zhu2022challenges, lillienskiold2022human} due to uncertainties in behavior during manipulation that require real-time feedback to drive the task to completion. 

Manipulation of limp, organic objects introduces additional complexities due to their inherent unpredictability and delicate nature and because their biological origin imposes natural variation in size, shape, surface, transparency, and rheological properties \cite{misimi2017robust}. 
In particular, advanced manipulation techniques - such as shape manipulation of elastoplastic volumetric objects (changing the shape of the object from an arbitrary initial to a desired shape) \cite{Billard2019TrendsManipulation, sivertsvik2024}, topology-altering state changes (slicing the object into smaller pieces) \cite{Billard2019TrendsManipulation} — are tasks that are still particularly challenging \cite{Billard2019TrendsManipulation}. 
Existing work largely focus on elastic objects \cite{thach2022learning, fonkoua2024deformation, qi2023adaptive} or non-volumetric objects such as cloth and ropes \cite{mitrano2021learning, shivakumar2023sgtm, miller2012geometric, matas2018sim, wu2019learning}.
Settings where the robot manipulates the object with a hand-held tool present its own set of challenges \cite{Billard2019TrendsManipulation, suresh2024neuralfeels, sun2021robotic}, particularly for tasks like shape manipulation or cutting. 
Moreover, bimanual and multi-arm manipulation \cite{Billard2019TrendsManipulation}, including the coordination of a second arm to enhance extrinsic dexterity, are areas that remain largely underdeveloped in robotics \cite{Billard2019TrendsManipulation}. 
These techniques are virtually nonexistent for the manipulation of natural deformable objects due to the inherent challenges the manipulation of such objects incurs. 
To master dexterous in-hand manipulation, the inclusion of touch perception is paramount \cite{suresh2024neuralfeels}.
Compared to other approaches that are purely vision-based, use traditional force control from sparse sensors \cite{xu2023, aly2024tactile}, or are predominantly developed for rigid objects \cite{chen2022visuotactile}, we consider an approach based on geometrical information from a high-resolution tactile sensor, analogous to how we humans detect haptic events in our hand when cutting with a knife. Shape servoing, volumetric deformable object manipulation, and the use of multiple arms to enhance dexterity are highlighted as major open challenges in robotic manipulation by \cite{Sanchez2018} and \cite{Herguedas2019}. Our system addresses these gaps through zero-shot non-prehensile shape servoing and volumetric manipulation of elastoplastic biological objects using a tri-manual setup, enabling precise and robust control in a fully autonomous framework.
Last but not least, tasks involving grasping with high precision accuracy \cite{sun2021robotic}, such as grasping limp objects, are particularly challenging since, on the one hand, one needs precise and consistent grasping. On the other, the grasping needs to be gentle so as not to destroy or degrade the quality of the product. 

\begin{table}[]
    \centering
    \setlength{\tabcolsep}{0.35em}
    \begin{tabular}{ lccccccccccccc }
    \toprule
     & \cite{Higashimori2010} & \cite{Matl2021} & \cite{Duan2024} & \cite{Bartsch2024} & \cite{Shi2023} & \cite{allison2024hashi} & \cite{aly2024tactile} & \cite{long2013modeling} & \cite{mason2022smart} & \cite{mason2024robutcher} & \cite{wright2024safely} & \cite{xu2023} & \textbf{Ours} \\
    \midrule
    Shape manipulation & $\checkmark$ & $\checkmark$ & $\checkmark$ & $\checkmark$ & $\checkmark$ & - & - & - & - & - & - & - & $\checkmark$ \\
    Cutting & - & - & - & - & $\checkmark$ & - & $\checkmark$ & $\checkmark$ & $\checkmark$ & $\checkmark$ & $\checkmark$ & $\checkmark$ & $\checkmark$ \\
    Grasping and picking & - & - & - & - & - & $\checkmark$ & - & $\checkmark$ & - & $\checkmark$ & - & - & $\checkmark$ \\
    \midrule
    Non-prehensile manipulation & - & $\checkmark$ & $\checkmark$ & - & $\checkmark$ & - & - & - & - & - & - & - & $\checkmark$ \\
    Tool switching & - & - & - & - & $\checkmark$ & - & - & - & - & - & - & - & $\checkmark$ \\
    In-hand tool manipulation & - & - & - & - & - & - & - & - & - & - & - & - & $\checkmark$ \\
    \midrule
    Visual perception & - & $\checkmark$ & $\checkmark$ & $\checkmark$ & $\checkmark$ & - & - & $\checkmark$ & $\checkmark$ & $\checkmark$ & $\checkmark$ & - & $\checkmark$ \\
    Tactile perception & $\checkmark$ & $\checkmark$ & - & - & - & $\checkmark$ & $\checkmark$ & $\checkmark$ & $\checkmark$ & $\checkmark$ & $\checkmark$ & $\checkmark$ & $\checkmark$ \\
    \midrule
    Learning-based  & - & $\checkmark$ & $\checkmark$ & $\checkmark$ & $\checkmark$ & - & - & - & $\checkmark$ & $\checkmark$ & $\checkmark$ & $\checkmark$ & $\checkmark$ \\
    Zero-shot sim-to-real & - & - & - & - & - & - & - & - & - & - & - & $\checkmark$ & $\checkmark$ \\
    \bottomrule
    \end{tabular}
    \caption{Features and capabilities of related work on limp object manipulation.}
    \label{tab:related-work}
\end{table}

Sashimi-Bot integrates these capabilities and features into a holistic robotic framework for preparing sashimi.
A comparison with studies on similar manipulation of natural limp objects is summarized in Table \ref{tab:related-work}.
Most work on shape manipulation of volumetric elastoplastic objects sculpt dough-like materials \cite{Higashimori2010, Matl2021, Duan2024, Bartsch2024, Shi2023}, typically through top-down pinches or non-prehensile techniques such as rolling or pressing.
While these modalities are appropriate for dough or clay, they may damage the delicate muscle tissue of salmon, so we instead opt for more gentle horizontal pushes. 
Work on meat handling and cutting systems largely focus on industrial settings \cite{xu2023robotization} and specialized hardware \cite{mason2022smart, allison2024hashi}, perhaps best exemplified by the Meat Factory Cell concept \cite{alvseike2018meat, long2013modeling, mason2024robutcher} or limit the scope to just the cutting task or sensor processing \cite{aly2024tactile, xu2023, wright2024safely}.
In contrast, our tri-manual system considers the full task and uses generic tools and control schemes that more closely approximate human-like manipulation.
In virtually all the referenced works, the tools used for shaping and cutting are manually fixed to the robot's end-effector. 
An exception to this is RoboCook \cite{Shi2023}, which automatically switches between tools during operation, but the instruments are all custom designed for the robot.
Instead of using affixed, high-tech tools, we show that good results can be achieved with general-purpose robotic arms, a simple hand-held pushing tool, a conventional hand-held chef's knife, and 1-DoF actuated wooden chopsticks. 
Machine learning techniques have proven valuable for processing rich visuo-tactile sensor data and dealing with the complexities of modeling and controlling deformable materials \cite{Matl2021, du2021high, Bartsch2024, Shi2023, mason2022smart, mason2024robutcher, wright2024safely, xu2023}.
However, with the exception of \cite{xu2023}, all references rely on expensive training data collected from a physical system.
In contrast, our shape manipulation controller transfers directly from simulated training to a real robot.
Lastly, where others assume known initial position or geometry \cite{Higashimori2010, mason2022smart, mason2024robutcher, aly2024tactile, Bartsch2024}, use teleoperation \cite{allison2024hashi}, or a human collaborator \cite{wright2024safely}, we present an autonomous system capable of handling and cutting limp, deformable salmon loins, regardless of initial position and shape.
Our system also constitutes an autonomous end-to-end pipeline - from initial shape manipulation and positioning, to cutting, and finally to placing the slices on a tray ready for serving - with the following contributions:

\begin{itemize}
    \item A non-prehensile shape servoing approach based on Deep Reinforcement Learning (DRL), with robust zero-shot sim-to-real generalization to objects with different physical properties, that gently brings the deformable salmon loin from an initial arbitrary configuration into a desired configuration suitable for cutting.
    \item A bi-manual, visually-controlled subsystem for cutting with an anthropomorphic robot hand wielding a conventional chef's knife, along with a motion planning algorithm for flexibly generating cutting trajectories. 
    \item A classification model for real-time processing of tactile feedback from a GelSight sensor placed on the knife blade that can detect when the knife touches the board and adjust the cutting trajectory accordingly. 
    \item A closed-loop flexible grasping framework based on visual servoing that is capable of delicately picking sashimi slices with a pair of chopsticks from multiple surfaces, including the knife itself, for the edge cases when slices are stuck after cutting.
\end{itemize}

We provide individual, quantitative ablation experiments for each component, as well as a holistic evaluation of the whole system autonomously preparing sashimi.

%% file: sec/2_system.tex
\section{Sashimi-Bot}
\label{sec:system}

\begin{figure}
\centering
\includegraphics[width=\textwidth]{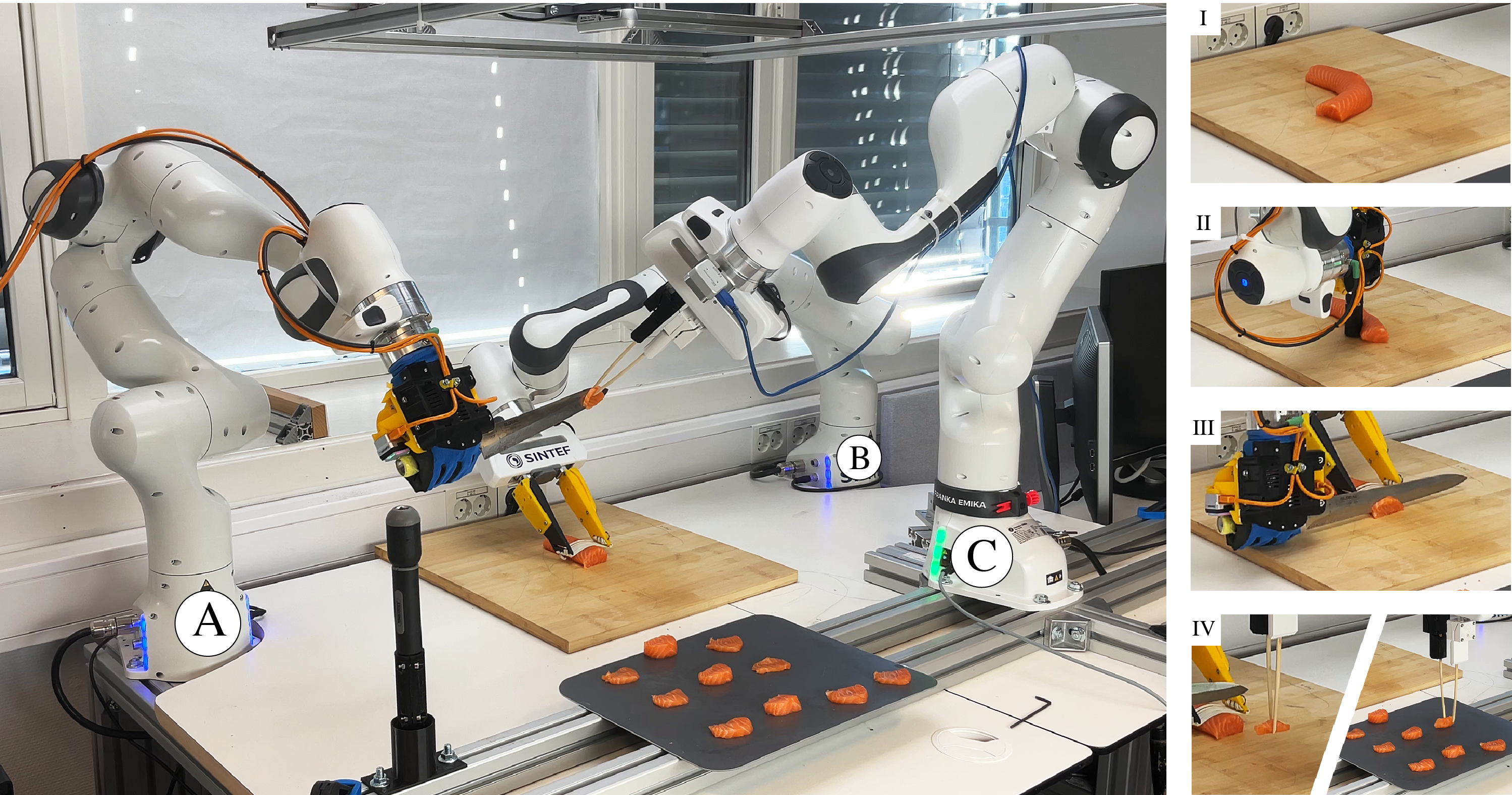}
\caption{Sashimi-Bot is a trimanual robotic framework capable of autonomously preparing sashimi. Left: the system in action. Right: starting with a salmon loin in an arbitrary initial configuration (I), the high-level components of the system involve shape manipulation (II), cutting (III), and picking and placing (IV).}
\label{fig:setup2}
\end{figure}

Sashimi-Bot consists of three 7-DoF robotic arms surrounding a shared working area (Figure \ref{fig:setup2} left).
The first robot (A) is equipped with a 1-DoF, compliant, antropomorphic gripper (QBSoftHand), which is used to grab auxiliary tools for manipulation and cutting, as well as a GelSight sensor that provide tactile feedback of the grasped object.
The second robot (B) is equipped with a specialized end-effector consisting of two long fingers and an elastic band that is used to gently stabilize the loin during cutting. 
These two robots are controlled in an eye-to-hand configuration from a static RGB-D camera mounted over the workspace.
The third robot (C) holds a pair of chopsticks, mounted on a 1-DoF parallel-jaw gripper, and is equipped with a wrist-mounted RGB-D camera.
The overall pipeline (Figure \ref{fig:setup2} right) starts with a salmon loin placed in an arbitrary pose on the cutting board (I). Using robot A, Sashimi-Bot begins by repositioning and reshaping the loin from its arbitrary shape to a centered and straightened shape configuration (II). 
It then grabs the knife and cuts a slice of the loin with robot A while stabilizing the object with robot B (III). 
Finally, robot C is engaged to pick up the newly cut sashimi slice and place it on a serving tray (IV). 
Steps III and IV are repeated until the entire loin has been processed or a predetermined number of cuts have been made.

\begin{figure}
\centering
\includegraphics[width=\textwidth]{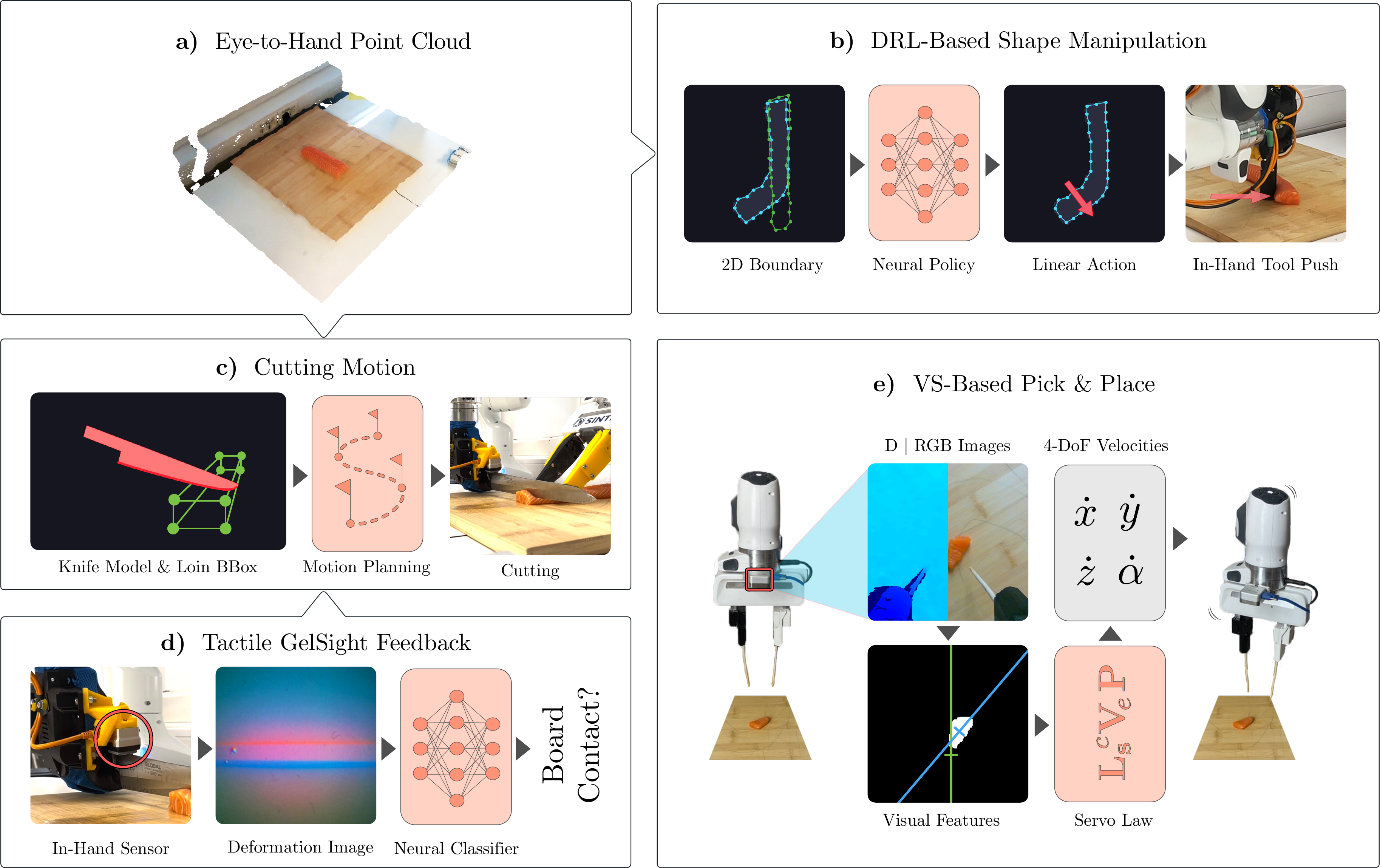}
\caption{
High-level components of the Sashimi-Bot pipeline.
\textbf{a)} A pointcloud of the workspace is provided by a static, top-down camera. For shape manipulation, we present \textbf{b)} a non-prehensile method based on Deep Reinforcement Learning (DRL).
\textbf{c)} Cutting trajectories are generated by a motion planning algorithm with \textbf{d)} tactile feedback from GelSight sensors. \textbf{e)} The pick-and-place module uses feedback from a separate, wrist-mounted camera and Visual Servoing (VS) to grasp sashimi pieces with a pair of chopsticks.
}
\label{fig:pipeline}
\end{figure}

An overview of the methods is given in Fig. \ref{fig:pipeline}. 
For shape manipulation, we employ a novel, non-prehensile approach based on Deep Reinforcement Learning (DRL, Fig. \ref{fig:pipeline}b).
The loin is driven towards a desired configuration using a series of discrete pushes with an in-hand manipulation tool. 
The non-prehensile approach allows for very gentle manipulation, as it circumvents the need of physically grasping the loin.

Loin shape configurations are represented as sets of image points sampled uniformly from the contours of segmentation images generated from a top-down projection of the workspace point cloud -- current (blue) and desired (green).
From there, the agent uses a neural policy based on the transformer
architecture~\cite{vaswani2017attention} to decide which contour point to push from and how far to push.
The policy is trained exclusively in simulation and deployed to the physical robot without any real-world training.

Once the loin is straightened and positioned, the cutting module assumes control of robot A and B (Fig. \ref{fig:pipeline}c).
Given the pose and bounding rectangle of the loin, estimated from the workspace pointcloud, a suitable cutting trajectory is generated with inverse kinematics \cite{9561366}. 
The motion planning algorithm moves the knife through the loin in one or more strokes while gradually lowering it towards the cutting board.
The resulting path is constrained so that, at any point in time, the tangent of the edge segment that is currently intersecting the loin's center is always parallel with the cutting board. 
Each cutting path is generated independently and can be configured with a desired slice thickness, how many strokes to use, and tilt of the knife blade (rotation around the axis running down the knife's spine).

Cutting with a conventional knife held in a soft gripper introduces uncertainty about the exact pose of the knife, and thus uncertainty about the exact moment it will hit the cutting board.
We present a solution to this issue by leveraging tactile feedback from a GelSight sensor (Fig. \ref{fig:pipeline} d). 
The sensor is mounted at the front of robot A's hand so that it makes contact with the blunt top-edge of the knife. 
Deformations on the gel pad of the sensor are captured as a video stream by an RGB camera positioned beneath the gel pad. 
We collect and annotate sensor data, and train a lightweight vision transformer classifier network to determine whether a sensor image corresponds to table contact. 
The trained model can then continuously classify tactile data during cutting, allowing us to adjust the planned trajectory if premature contact is detected.

The pick-and-place module coordinates with the cutting controller to locate, pick, and place newly cut slices onto a serving tray (Fig \ref{fig:pipeline} e).
Immediately following a cut, a search region (e.g. on the cutting board or the knife's blade) is sent to robot C, which then moves its end-effector into position.
From there, feedback from the wrist-mounted RGB-D camera is used to control the arm with a closed-loop control law based on visual servoing.
We found the closed-loop approach advantageous as it provides robustness to calibration errors and sensor noise. 
The orientation of the plane on which the slices lie is known: it is either horizontal if the slice lies on the cutting board, or available from kinematics of robot~A if the slice is on the knife’s blade, so only the 4 remaining DoF have to be controlled. For that, the sashimi slice is located by segmenting the image with an HSV band-pass filter, and visual features are generated based on the 3D centroid and principal axis of the foreground pixels.
In addition to their thematic fit, the wooden chopsticks serve a practical purpose; chopstick-based grippers have previously been shown to work well for delicate picking \cite{allison2024hashi} and we found their natural compliance and slender profile well-suited for gently picking up the thin sashimi slices.

%% file: sec/3_results.tex
\section{Results}
\label{sec:results} 

We provide individual experiments for each component of Sashimi-Bot in isolation, as well as a holistic evaluation of a fully autonomous sashimi preparation pipeline, including shape manipulation, cutting, and picking.

\subsection{DRL-Based Shape Manipulation}
\label{sec:result-drl-shape-manip}

\begin{figure}[ht]
    \centering
    \includegraphics[width=\textwidth]{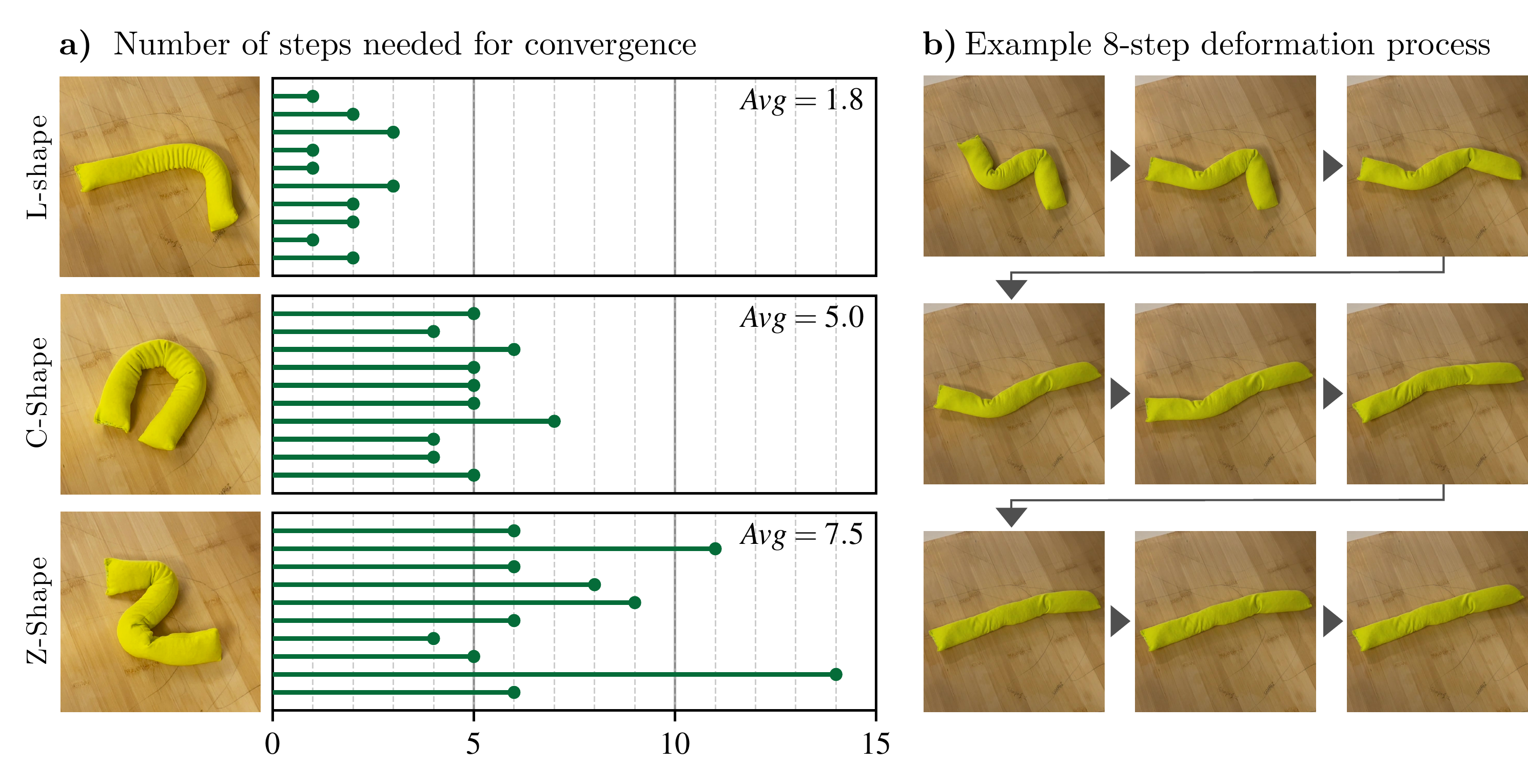}
    \caption{DRL-based shape manipulation results. \textbf{a)} Number of manipulation actions (steps) needed to reach convergence for each initial shape, \textit{L},  \textit{C}, \textit{Z}, across 10 randomized trials. \textbf{b)} Sample straightening of a Z shaped object.}
    \label{fig:drl-result}
\end{figure}

After training the DRL-based shape manipulation policy in simulation, we deploy it directly to the real world and validate the controller with a rice-filled cloth object.
To assess performance, we measure the number of pushing actions needed to straighten the object.
Using three different initial shapes (Fig. \ref{fig:drl-result}), we run ten trials for a total of 30 shape manipulation problems.

We observe very robust performance. In all trials, the DRL controller successfully repositions and straightens the object (Fig. \ref{fig:drl-result} a).
For the \textit{L-Shape}, the easiest shape, it used 1.8 actions on average; for the slightly harder \textit{C-Shape} shape, the average was 5.0 actions; and for \textit{Z-Shape}, the average was 7.5 actions (Fig. \ref{fig:drl-result}a). 
We note that in many trials, it got very close in only a few moves and then spent a longer time making fine adjustments before terminating (e.g. Fig. \ref{fig:drl-result}b).
In a few cases, we observed that the controller overshot by generating a push action that was too long, but its closed-loop nature allowed it to correct quickly.

\subsection{Cutting Motion}
\label{sec:results_cutting}

\begin{figure}[ht]
    \centering
    \includegraphics[width=1.0\textwidth]{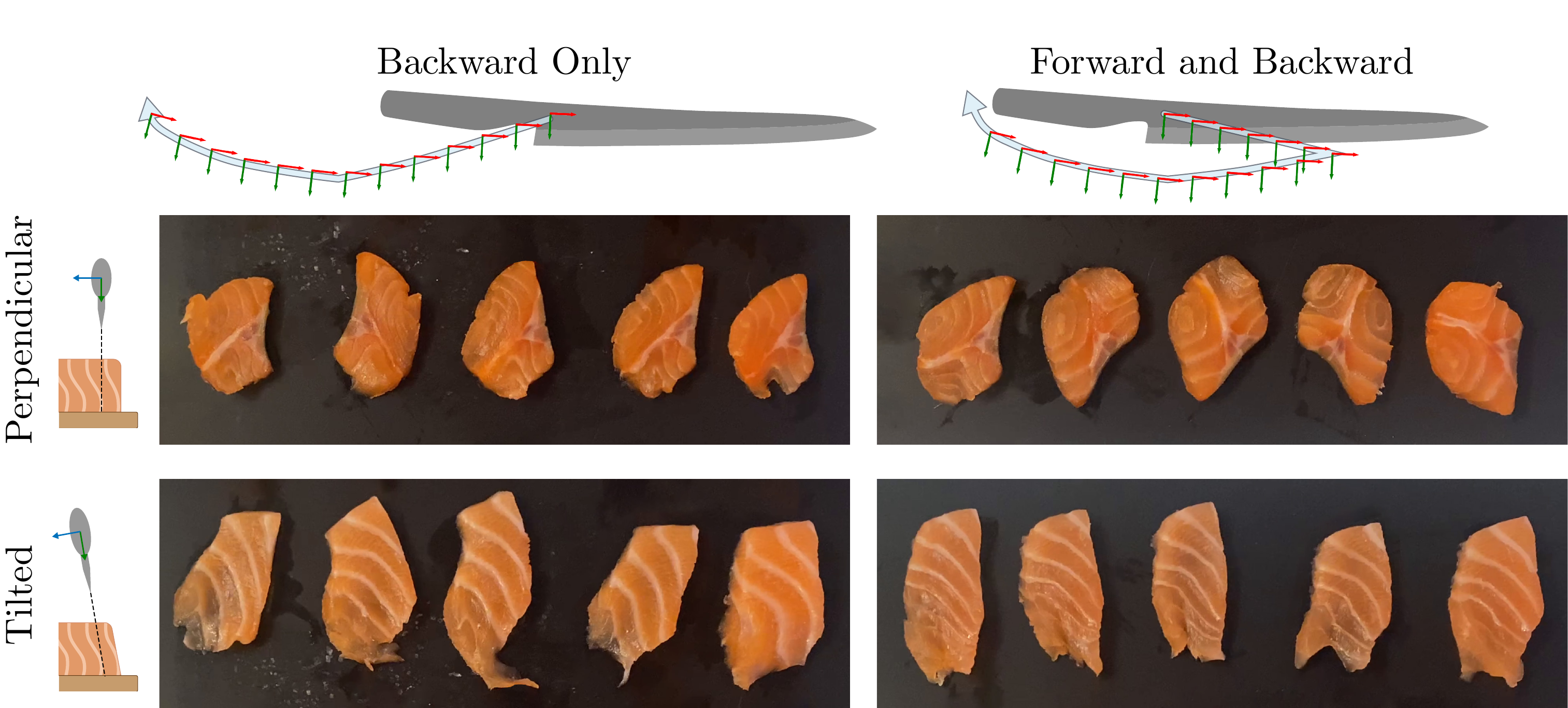}
    \caption{Cutting results across different configurations. We vary whether to cut with the knife perpendicular to the cutting board or tilted ($\sim20$ degrees), and whether to cut with a single \textit{backward} stroke or dual \textit{forward-backward} strokes}
    \label{fig:cutting-results}
\end{figure}

The cutting controller is evaluated in four different configurations corresponding to variation in tilt angle and the number of cutting strokes (Fig. \ref{fig:cutting-results}).
In preliminary experiments, we found the initial cut more challenging than subsequent ones due to a combination of sensor noise and the lack of a clean edge at the end of a natural loin.
Additionally, thinner cuts are harder than thicker ones, with degrading performance observed for thicknesses $<5$~mm.
In Fig. \ref{fig:cutting-results}, each cutting mode is evaluated with five (non-initial) cuts of 7 mm thickness. 
All cuts have successful separation and we observe that the slices have smooth surfaces, uniform textures, and a consistent thickness, indicating that all four modes yield consistent cutting performance. The observed difference in size is mainly due to natural variation in the cross section of the loin, although tilting the blade produces silghtly wider slices. Breaking the cut into multiple strokes allows for a longer total cutting motion, which is useful if the knife is relatively short compared to the height of the object being cut. 

\subsection{Tactile Feedback} 
\label{sec:results-tactile}

\begin{figure}
    \centering
    \includegraphics[width=1.0\linewidth]{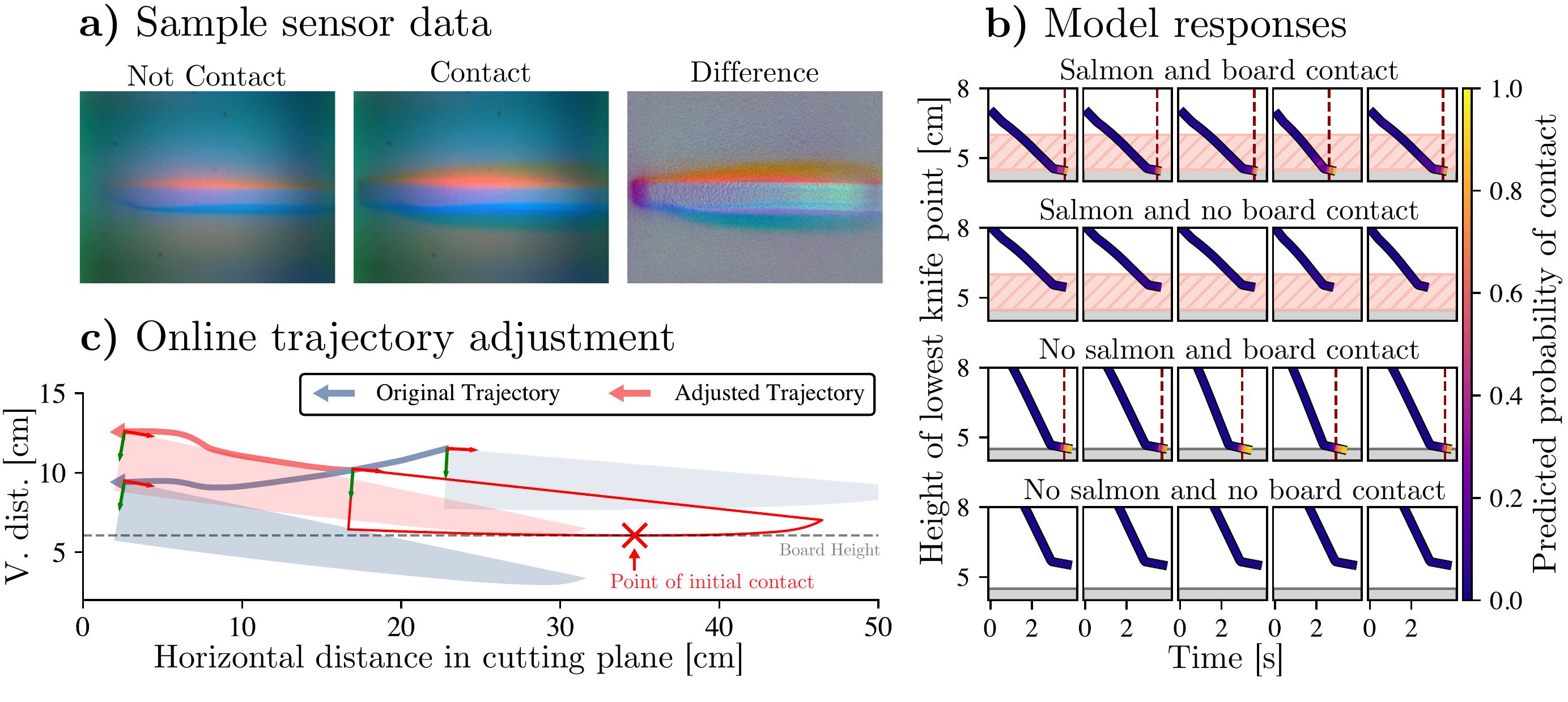}
    \caption{Tactile feedback results. \textbf{a)} Sample GelSight sensor images from non-contact and contact with the cutting board, and the difference between the two. \textbf{b)} Predicted probability of board contact in four different scenarios. Dashed vertical line indicates first point in time where $\text{p(board)} \geq 0.5$. The first row shows model response for regular cuts through a salmon loin that end up in contact with the board, the second row shows cutting paths that only go through the salmon without hitting the board, the third and forth rows repeat these experiments without a salmon loin (cutting air). Note that contact is only detected when the knife actually hits the cutting board. \textbf{c)} Example of closed-loop trajectory adjustments based on tactile feedback. We raise the board height to provoke early contact with the knife. When contact is detected, the remaining trajectory is adjusted so that no point on the knife will ever go below the vertical position of the contact point. See the end of Section \ref{sec:gelsight-feedback-appendix} for more details.}
    \label{fig:gelsight-results}
\end{figure}

We train a real-time classification model that can detect, based on GelSight images, when the knife touches the board
and adjust the cutting trajectory accordingly. For this purpose, we collect and annotate a dataset of 12,397 GelSight sensor readings corresponding to 157 cutting trajectories and train a lightweight, transformer-based classifier.
An example of the sensor data can be seen in Fig. \ref{fig:gelsight-results}a.
On a held-out validation set of 20 cutting trajectories, we obtain an accuracy of 95\%, a precision of 99\%, and a recall of 67\%.  
The precision-recall discrepancy is a product of training data class imbalance, and that readings immediately following contact are virtually indistinguishable from non-contact ones.
However, the model is highly reliable for non-ambiguous contact events.
To demonstrate this, we collect an additional 20 cutting trajectories in four different scenarios (Fig.~\ref{fig:gelsight-results}b).
In the first two, we cut through a salmon loin, but only half goes through to the board. 
We observe that the model successfully detects board contact (red dotted line) in (and only in) the relevant trajectories.
Moreover, the modeled probability of contact stays close to zero when there is no contact and smoothly increases as the knife is pressed into the cutting board.
Repeating the experiment without a salmon loin yields consistent results.

The model's feedback can be used to close the loop of the cutting controller.
When contact with the board surface is detected, we calculate the point of contact, which is given by the lowest point on the knife model in the current end effector pose. 
The remaining cutting trajectory is adjusted with vertical displacements such that no point on the knife will ever go below the point of contact, an example of which is shown in Fig. \ref{fig:gelsight-results}c.
This closed-loop approach may provide robustness to slight variations in the knife grasp and allows reliably cutting all the way through the loin by deliberately configuring the initial planned trajectory to intersect the cutting board with some margin.

\subsection{Vision-Based Picking}
\label{sec:results-picking}

We evaluate the 4DoF (position and orientation) closed-loop
vision-based pick-and-place 
controller with visual features from sashimi slices of salmon, pork, and cod, as illustrated in Fig. 6. The slices are grabbed from the sides, squeezed inwards and lifted to be placed on the tray.
The pork meat is generally firmer and more uniform than the salmon, whereas the cod is more fragile and consists of chunks of meat held together by thin ligaments.
We use five different slices of each meat and pick each slice up from a randomized pose five times, for a total of 75 trials.
The controller parameters are held constant across the entire experiment, except for the color-based segmentation system calibrated for each meat type in order to generate the object mask.
A picking trial is successful if the sashimi slice is lifted from the board and placed on a tray.
The controller is successful in 74/75 trials, the one failure being due to an error in the segmentation system caused by a shadow cast by an oblique light source.
The system is also very gentle. Despite being picked up several times, none of the slices tore or were otherwise damaged during the experiments.

\begin{figure}
    \centering
    \includegraphics[width=1.0\linewidth]{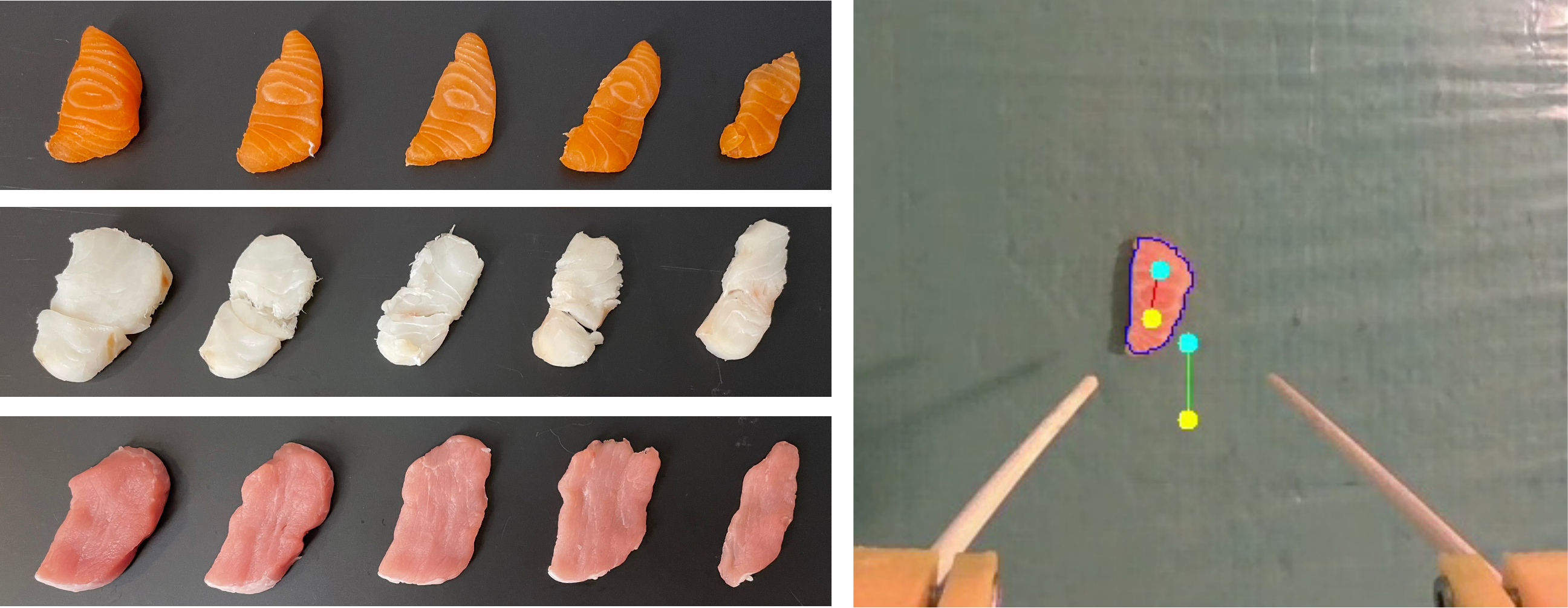}
    \caption{Picking experiments. Left: the slices of salmon, cod, and pork used during the picking experiments. Right: sample footage from the picking robot's wrist-mounted camera used to control grasping in the end-effector frame. The blue boundary shows the segmentation mask, the computed visual features are overlaid on the slice with two dots and a connecting red line to mark the principal axis of the object, whereas its desired value is shown in the center of the image with a green connecting line. The chopsticks are servoed until the principal axis (position and orientation angle) of current  and desired poses are aligned.}
    \label{fig:picking-results}
\end{figure}

\subsection{Full System}

We conclude the experiments by providing an end-to-end evaluation of a fully autonomous sashimi-cutting pipeline, including shape manipulation, cutting and picking.
The execution of the cutting and picking controllers is interleaved with a slight overlap so that a new slice is cut while the previous one is being placed on the tray.  
A recurring edge case discovered during preliminary experiments was that the sashimi slice would occasionally get stuck to the knife's blade, particularly during thin cuts with a tilted knife.
We resolve this issue by picking the sashimi directly from the knife.
If the picking robot (C) cannot find the sashimi slice in its initial search region, it will report back to the cutting robot (A), which in turn will present the knife and send a new search region placed along the blade.

We experiment with the same three initial shapes as shown in Figure \ref{fig:drl-result}, as well as an arbitrary shape obtained by throwing the loin onto the cutting board to land in an arbitrary initial shape.
A salmon loin is wetter and heavier than the manipulation test object used in previous experiments, leading to substantially larger friction with the wooden cutting board.
This poses a challenge for our push-oriented manipulation scheme, as it is difficult to completely smoothen out wrinkles if the loin gets compressed along its major axis.
Despite the change in dynamics, the controller managed to adequately reposition and straighten out the loin from all initial configurations, completely avoiding the need for manual intervention prior to cutting. 
This demonstrates that our shape manipulation method generalizes well, not only from simulation to the real world, but also across deformable materials with varying compliance, texture, and rheological properties.
Across the diffferent initial shapes, the controller used between 1 and 4 manipulation actions, with each pushing action taking betwen 4 and 8 seconds to execute (depending on start and end positions -- average 6.1 sec).

We cut a total of 34 sashimi slices with a forward-backward stroke while varying cutting thickness (6-16 mm) and knife tilt angle (0-20 degrees).
To facilitate picking, an extra segment was appended to the planned cutting motion, which pushes freshly cut slices approximately 9 cm away from the main loin in a translational swipe to the right.
Across the 34 slices, six got stuck to the knife blade, and 28 remained on the cutting board.
Among the former six, all were successfully picked from the knife.
Among the latter 28, the robot successfully picked up 26 from the cutting board.
The two failures were caused by slips when picking up very thin slices. 
A cut takes on average 5.3 seconds, but the cutting robot has to wait for the picking robot to lift up the slice before it can start on a new one, which puts average time between consecutive cuts at 27.9 seconds for normal picks and 37.7 seconds when picking from the knife blade. We report time metrics for completeness, but note that the focus of this paper is robust autonomy, not speed.

%% file: sec/4_discussion.tex
\section{Discussion}
\label{sec:discussion}

Autonomously preparing sashimi is an interesting microcosm of several open problems within robotics.
Success is dependent on the collaboration of several subsystems, including visual and tactile perception, coordination of multiple robots, manipulation of elastoplastic objects, and tool usage.
We have presented Sashimi-Bot, a tri-manual robotic framework capable of autonomously manipulating, slicing, and serving salmon loins.
Similarly to humans, our system leverages intuition about deformable object dynamics and employs visual and tactile sensory feedback to control manipulator arms.
The framework is validated empirically with experiments of each subsystem, as well as an integrated test of the entire pipeline.

Further improvements will likely be realizable by incorporating richer haptic (tactile and force) and visual feedback, as well as active perception.
While a cornerstone of the shape manipulation and cutting pipelines, we observed the static top-down camera to be a limiting factor for both tasks.
Specifically, moving manipulator arms and tools above the cutting board would often occlude the loin, complicating smooth closed-loop control. 
Additionally, capturing the entire working area from a single stationary camera inherently leads to resolution trade-offs, compromising the precision in downstream tasks.
While both of these problems could, in principle, be solved by utilizing more cameras, this introduces a computational burden that may compromise the real-time constraints. A more interesting avenue of research and a more efficient solution is to incorporate active vision.
In particular, we hypothesize that using only the picking robot's wrist-mounted camera and an intelligent, active perception strategy could greatly diminish the aforementioned limitations. 
High-resolution visual feedback of the knife from an up-close perspective could also be used in conjunction with richer tactile feedback and joint force feedback to allow for precise online estimation of its relative transform to the gripper and to more tightly close the loop, allowing for more sophisticated and precise cutting trajectories. Additionally, this strategy would also enable cases involving dynamic scenarios where the object moves during cutting or when one needs an online adjustment of the knife's 6DoF-pose to compensate for the deformation of the object during the previous cutting iteration.

Despite these limitations, we believe that Sashimi-Bot represents a milestone in advanced robotic manipulation and cutting of deformable, volumetric objects. 
We hope this work inspires others and facilitates real-world applications and deployments of robots to a broad range of applications involving deformable objects in general, and limp muscle food objects in particular, that are well known to be challenging during robotic manipulation.

We envisage that the implications and the impact of the robotic manipulation technology our system represents may be broad, with the potential to unlock a transformation capable of addressing several pressing global issues related to sustainable development \cite{guenat2022sustainable}. If we are to ensure global food security in a long-term perspective, given the pressing need for sustainable protein sources, and address the environmental challenges of food production \cite{lipson2023robots}, advancing robotic manipulation is crucial \cite{hodson2017food}. In addition, robotic manipulation of natural deformable objects enhances resilience during crises, such as logistics chain disruptions or lockdowns, when human access to plants is limited \cite{isachsen2021gpu}. Seafood is one of the most sustainable protein sources \cite{Bianchi2022}, offering a low environmental impact compared to other animal-based foods. Currently, on a global scale, only 17\% of the protein sources come from seafood \cite{costello2020future}. However, to meet the increasing global demand for protein \cite{costello2020future}, a breakthrough in robotic deformable object manipulation technology is imperative \cite{zhu2022challenges}, yet such technology is currently nonexistent for advanced robotic manipulation of challenging natural objects \cite{sivertsvik2024}, such as the exemplar objects used in our system. Salmon and other similar natural deformable objects present some of the greatest challenges in manipulation due to their fragility, variability, and uncertainty in behavior during manipulation \cite{misimi2016gribbot, lillienskiold2022human}. The current lack of advanced robotic technology forces producers to transport seafood to centralized processing facilities or to low-cost countries overseas, resulting in inefficiencies, higher costs, and significant carbon emissions due to transport and retransport. By endowing robots with the capability of precise and advanced manipulation of natural deformable objects, as represented by Sashimi-Bot, we can unlock transformative solutions that allow for food processing closer to harvesting sites, reducing transportation needs, preserving food quality, and increasing production capacity \cite{guenat2022sustainable}. Such a robotic technological advancement has the potential to contribute to ensuring food security \cite{hodson2017food} and food self-sufficiency, reducing environmental impacts, and revolutionizing sustainable food and protein production for the future \cite{guenat2022sustainable}, as well as for other real-world applications.
Principles may transfer to other domains, such as soft tissue handling and cutting in surgery, processing textiles and leathers for tailoring applications or other elastoplastic objects in manufacturing, and pruning and harvesting in agricultural and horticultural settings.
While hyper-specialized hardware is often the way to maximize efficiency, more control and learning-oriented  approaches that can utilize existing tools made for humans - like Sashimi-Bot - may have an advantage in fast-paced and changing domains where constant retooling is time-consuming or costly.

\section*{Funding Declaration}
This work was supported by the Research Council of Norway under project grants GentleMAN (299757) and BIFROST (313870).

%% file: sec/5_content.tex
\section*{Data Availability}

Source Data used to generate summary statistics and plots are provided with this paper. A complete repository of relevant data and media will be made available upon publication in a public repository compliant with the Norwegian Government Authority policies for data sharing\footnote{\url{https://www.regjeringen.no/en/dokumenter/national-strategy-on-access-to- and-sharing-of-research-data/id2582412/?ch=2}}.
\\
\section*{Code availability}
Based on the Norwegian Authority for Export Control\footnote{\url{https://www.regjeringen.no/en/topics/foreign-affairs/export- control/id754301/}} to safeguard against potential misuse and existing sanctions related to some of the countries involved currently in conflicts, the full source code associated with this research will not be made publicly available. However, relevant source code for the core contributions has been provided for the editors and reviewers. Additional pseudo code, always in compliance with the Export Control Authority, can be made available upon request.

\section*{Author Contributions}
S.H. contributed to the main ideas and the experimental design,
implemented the shape manipulation, grasping, and contributed to
implementation of tactile feedback, contributed to the full system integration
and evaluation, performed the experiments and data analysis, and wrote the
paper. A. P. contributed to the main ideas, formulation and implementation
of the tactile feedback, performed the experiments, writing of the
paper. E.R.Ø. contributed to the main ideas, performed the experiments,
system integration and evaluation. F.Z. contributed to the main ideas,
formulation and implementation of the cutting, performed the experiments,
writing of the paper. F.M. contributed to the main ideas, performed the
experiments and data analysis, system integration. A.L. contributed to the
main ideas, multi-robot system integration and tools, performed the
experiments and system evaluation. A.P.S. contributed to the main ideas,
grasping, writing the paper. E.H.A. contributed to the main ideas, formulation
and implementation of the tactile feedback. F. C. contributed to the main
ideas, experimental design, formulation of grasping, writing of the paper. A.K.
contributed to the main ideas, experimental design, writing of the paper. P.C.
contributed to the main ideas, formulation and implementation of the cutting,
writing of the paper. E.M. contributed to the main ideas and did the
formulation of the experimental design, analysis of experiments, experiments
and full system evaluation, writing the paper, and provided funding.

%% file: app/1_methods.tex
\section{Methods}
\label{app:methods}

\input{app/1a_rl_based_shape_manipulation}
\input{app/1b_cutting_motion}
\input{app/1c_gelsight_feedback}
\input{app/1d_vs_based_grasping}

%% file: app/1a_rl_based_shape_manipulation.tex
\subsection{RL-Based Shape Manipulation}
\label{sec:method-rl-based-manipulation}

The RL-based shape manipulation controller drives the object toward a desired configuration non-prehensilely with a sequence of discrete push actions. 
The problem is modeled as a goal-conditioned Markov Decision Process (MDP), with observations $o \in \mathcal{O}$, goals $g \in \mathcal{G}$ , actions $a \in \mathcal{A}$, and rewards $r \in \mathcal{R}$.
Both the observations $o$ and goals $g$ are given as binary segmentation images $\in \{0, 1\}^{H \times W}$ of the object from a top-down perspective, with $o$ corresponding to the current configuration and $g$ being a pre-configured goal configuration.
The agent picks actions $a$ which parametrizes the starting position, direction, and extent of a linear push in the image plane. 
After each action, the agent receives an observation $o'$ of the successor state, as well as a reward~$r$ proportional to the Intersection over Union (IoU) between $o'$ and $g$.

\begin{equation}
    r(o, g, a, o') = r(g, o') = \min_{\mathcal{T} \in \mathcal{N}} \frac{\sum_{i=1, j=1}^{H \times W} \mathbbold{1}\left[o'(i, j) \land \mathcal{T}(g)(i, j)\right]}{\sum_{i=1, j=1}^{H \times W} \mathbbold{1}\left[o'(i, j) \lor \mathcal{T}(g)(i, j)\right]}   
    \label{eq:rl-reward}
\end{equation}

Here, $\mathbbold{1}\left[\cdot \right]$ is an indicator function that evaluates to 1 if the predicate inside is true and 0 otherwise. The images are treated as boolean functions $H \times W \rightarrow \{F, T\}$ that evaluate to true if a pixel belongs to the foreground. $\mathcal{T}$ is a translation operator that shifts the image coordinates. We search for translations in $\mathcal{N}$, a $11 \times 11$ neighborhood centered on the identity operator, and pick the one yielding the lowest IoU. This makes the reward invariant to small translational discrepancies between $o$ and $g$, which allows the agent to prioritize the shape of the object as long as the overall position is approximately aligned.

The RL agent learns the controller with the Soft Actor-Critic (SAC) algorithm \cite{haarnoja2018soft}. This learning scheme consists of two neural networks: an actor policy network and a critic Q-function network. The actor policy, with parameters $\theta$, generates distributions $\pi_\theta(a | o, g)$ over the action space to determine actions based on observations and goals. The critic implements a Q-function $Q_\phi(o, g, a)$ with parameters $\phi$, which estimates the expected sum of future rewards after committing to an action, with future rewards discounted by factor $\gamma$. The system also includes an intrinsic, entropy-based reward that is added to the task reward. This term is calculated from the approximate policy entropy $\mathbb{E}[-\log \pi_\theta(a | o, g)]$ and scaled by weight $\alpha$, which helps maintaining diversity in the action selection. The two networks are trained with ($o, g, a, r, o'$) tuples sampled from a replay buffer that is gradually filled by rolling out the current version of the policy. The network parameters are updated with gradient-based optimization of the following objectives.

\begin{equation}
    \begin{split}
    J_\pi(\theta) &=\underset{a \sim \pi}{\mathbb{E}} \left[\min_{i \in \{1, 2\}}{Q_{\phi^i}(o, g, a)}  \right] - \alpha \log(\pi_{\theta}(a \mid o, g)) \\
    J_Q(\phi_i) &= \mathcal{L}\left(Q_{\phi_i}(o, g, a), r + \gamma \underset{{a' \sim \pi }}{\mathbb{E}}\left[\min_{j \in {1, 2}} Q_{\bar{\phi_j}}(o', g, a') - \alpha \log \pi_s(a'|o', g) \right]\right)
    \end{split}
    \label{eq:rl-objective}
\end{equation}

The actor is trained to maximize $J_\pi(\theta)$, which amounts to a balance of maximizing the Q-function and the immediate entropy bonus. 
The Q-function is trained with temporal difference learning \cite{sutton2018reinforcement} to minimize $J_Q(\phi_i)$, which encodes the discrepancy between predicted action-value $Q(o, g, a)$ and a one-step bootstrap estimate based on empirical reward $r$ and the estimated expected value of the successor state $o'$. In the expression above, $\mathcal{L}$ denotes a discrepancy loss function, in our case the Huber loss \cite{huberloss}. Similarly to SAC \cite{haarnoja2018soft}, we train two sets of Q-networks, denoted $\phi_1$ and $\phi_2$, and maintain exponential averages of them, denoted $\bar{\phi}_1$ and $\bar{\phi}_2$, to yield more stable bootstrap estimates. The entropy weight coefficient $\alpha$ is continuously adjusted to target a fixed entropy setpoint.

\begin{figure}
    \centering
    \includegraphics[width=\textwidth]{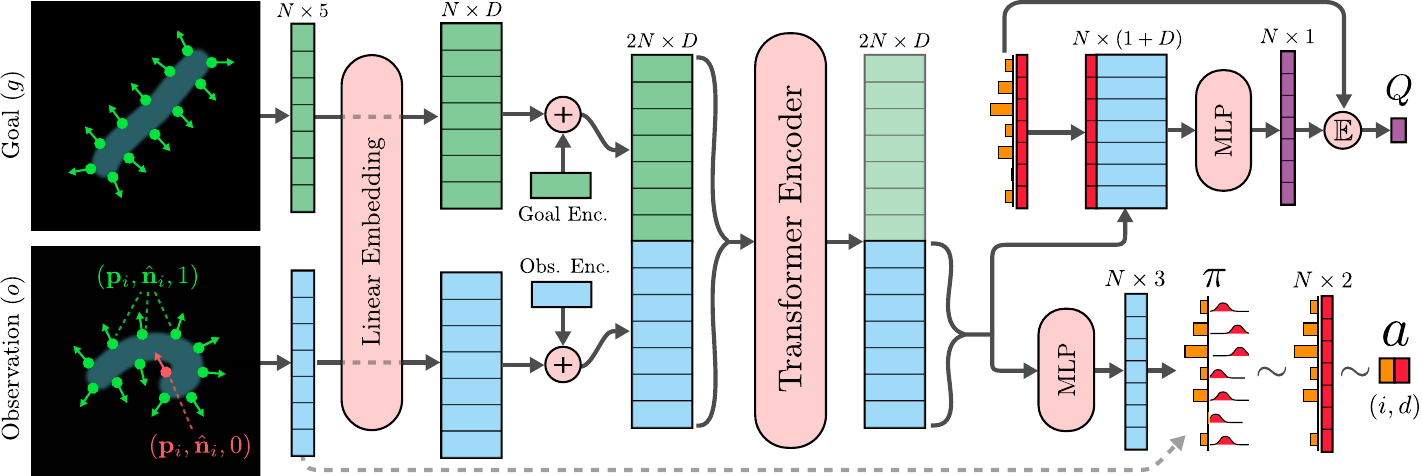}
    \caption{
    The neural network architecture $\pi(a | o, g)$ and $Q(o, g, a)$ in RL-based shape manipulation. From the left, we begin by uniformely sampling $N$ samples from the boundaries of segmentation masks $o$ and $g$. Each sample $i$ is described by its image coordinates $\mathbf{p}_i \in \mathbb{R}^2$, the boundary normal $\mathbf{\hat{n}}_i \in \mathbb{R}^2$, and a flag $v_i \in \{0, 1\}$ indicating whether there is space to initiate a push at point $i$. An affine map $\mathbb{R}^5 \rightarrow \mathbb{R}^D$ transforms each sample to $D$-dimensional vectors. To make the goal and observation vectors distinguishable, learnable encoding vectors are added in prior to concatenation.
    We use a transformer encoder \cite{vaswani2017attention} to mix all $2N$ samples from $o$ and $g$ but keep only the output corresponding to the $N$ tokens from $o$. 
    From here, different network heads are employed by the actor and the critic. 
    The actor models $\pi(a | o, g)$.
    The action $a$ consists of two parts $(i, d) \in \mathbb{N} \times \mathbb{R}$; respectively the index of the push origin $\mathbf{p}_i$ and how far to push along $\mathbf{\hat{n}}_i$.
    We model $p(i, d)$ with a mixture distribution factored into $p(i)$ and $p(d | i)$. 
    Here, $p(i)$ is a categorical distribution generated with a softmax (with zero probability for tokens with $v=0$), and $p(d | i)$ consists of $N$ gaussian distributions transformed by tanh and affine bijectors.
    The sufficient statistics for $\pi$ is given by an $\mathbb{R}^{N \times 3}$ matrix, which is computed with a Multilayer Perceptron (MLP). 
    After sampling $(i, d)$, the push action is given as $(\mathbf{p}_i, d \mathbf{\hat{n}}_i)$.
    The critic estimates value given $(o, g)$ and a half-sampled action $\{p(i), d_i\}_{i=1}^N$.
    It first uses an MLP to map the transformer output and $d$ to Q-values. It then calculates a single Q value by taking the expectation over $p(i)$. If a fully sampled action $(i, d) \in \mathbb{N} \times \mathbb{R}$ is provided (i.e. when sampled from replay buffer), $p(i)$ is constructed as a one-hot vector.
    }
    \label{fig:boundary-transformer}
\end{figure}

Crucially, neither $\pi_\theta$ nor $Q_\phi$ acts directly on the raw pixels of $o$ and $g$. 
Instead, they process a sparser representation of the object's boundary with a transformer-based architecture (Fig. \ref{fig:boundary-transformer}).
Preliminary experiments revealed this to lead to substantially more data-efficient learning than alternatives working directly on dense pixel representations. 
It also provides a convenient mechanism for constraining pushing actions to always originate outside the boundary and move directly into the object. 
The former is important to avoid the robot squishing the object from above when approaching.
The latter facilitates better sim-to-real generalization, as it prevents glancing pushes where the manipulation tool is dragged along the object boundary, a modality with significantly more complex contact dynamics that are difficult to get right in simulation.

The RL system is implemented in PyTorch \cite{paszke2019pytorch}. 
Each observation and goal is represented with $N=32$ boundary tokens and the Transformer Encoder consists of three $D=256$ dimensional unmasked self-attention blocks \cite{vaswani2017attention}. 
The MLPs for both the actor and the critic use the LeakyReLU activation \cite{maas2013rectifier} and a single hidden layer with 256 units.  
Training is carried out exclusively in a soft-body simulator based on Isaac Gym with the flex backend \cite{isaac}. 
The flex engine is a GPU-accelerated, particle-based physics engine, largely based on position-based dynamics \cite{macklin2014unified, muller2007position}, that can model both soft and rigid objects. 
We use a total of 8 rectangular soft-body meshes of varying scale and aspect ratio that resemble the geometry of a real salmon loin. 
The parameters governing the stiffness, damping, and friction of the soft body is empirically hand-tuned to visually match footage of its real-world counterpart undergoing manipulation.
Each episode is initialized by sampling an initial and desired configuration, given by a random translation and rotation of the soft-body in its natural (straightened) configuration. The robot arm is completely abstracted from the simulated environment and replaced by just the cylindrical pushing tool it wields in the real world.

\begin{figure}[htb]
    \centering
    \includegraphics[width=0.8\linewidth]{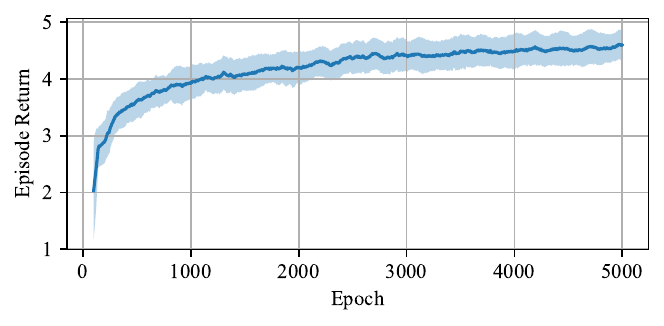}
    \caption{On-policy average episode return (sum of rewards (Equation \ref{eq:rl-reward}) across length-8 episodes) as function of training epoch with a smoothing window of 100 epochs. Shaded area indicate standard deviation. The agent is still marginally improving at the end of training, but we found that additional training led to be negligible improvement in overall system efficacy.}
    \label{fig:episode-return}
\end{figure}

We train on a server with a 40-core Intel Xeon Gold 6148 processor, 754 GB of system memory, and a 32GB NVIDIA V100 GPU. However, we note that less than 32 GB of system memory and less than 4 GB of GPU memory is ever allocated. A full training run involves sampling a total of 96,000 length-8 episodes, collected in sets of 32 and takes 40 hours on the given hardware. The episode reward as function of training epoch for the run that was used in the experiments is shown in Fig. \ref{fig:episode-return}.
Between each set, we update the actor and critic parameters 256 times with mini-batches of size 32. 
The live Q-networks parameters ($Q_{\phi_1}$ and $Q_{\phi_2}$) are updated first with the Adam optimizer \cite{kingma2014adam} and a learning rate of 0.0001, the actor ($\pi_\theta$) second with the same optimizer and a learning rate of 0.0003, and the target Q-networks ($Q_{\bar{\phi_1}}$ and $Q_{\bar{\phi_2}}$) last with an exponential averaging weight of $0.005$.  We use a discount rate of $\gamma = 0.95$ and a target entropy setpoint of $\mathbb{E}[-\log \pi_\theta(a | o, g)] = -1$.

We deploy the agent directly, without any real-world fine-tuning, as a server that the robot controller can query for action primitives. 
After being configured with an image $g$ of the desired object shape configuration, subsequent requests map observations images $o$ to push actions $a$ until convergence.
To automatically determine convergence, we initially experimented with the IoU used to calculate rewards. However, for the straight goal shapes considered in this work, we found the following criterion to work better. 

\begin{equation}
    \begin{split}
    Converged(o, g) &= 
        || \mathbf{\mu}(g) - \mathbf{\mu}(o)||_2 < \epsilon^{\text{SM}}_1
        \; \wedge \;
        \sigma^2_{\mathbf{\hat{e}}_{min}}\text(o) / \sigma^2_{\mathbf{\hat{e}}_{min}}\text(g) \leq \epsilon^{\text{SM}}_2 \\
    \mathbf{\mu}(x) &= \frac{
        \sum_{i=1, j=1}^{H \times W} \mathbbold{1}\left[x(i, j) \right] \begin{pmatrix} i \\ j \end{pmatrix}
    }{
        \sum_{i=1, j=1}^{H \times W} \mathbbold{1}\left[x(i, j) \right]
    }
    \\
    \sigma^2_{\mathbf{\hat{e}}_{min}}(x) &= \frac{
        \sum_{i=1, j=1}^{H \times W} \mathbbold{1}\left[x(i, j) \right]  \langle \begin{pmatrix} i \\ j \end{pmatrix} - \mathbf{\mu}(x), \mathbf{\hat{e}}_{min} \rangle^2
    }{
        \sum_{i=1, j=1}^{H \times W} \mathbbold{1}\left[x(i, j) \right]
    }
    \end{split}
    \label{eq:rl-manip-termination-criterion}
\end{equation}

Here, $\mathbf{\hat{e}}_{min}$ refers to the minor axis of the foreground pixels in $g$. The utility function $\mathbf{\mu}(x)$ computes the center of mass of the foreground object and $\sigma^2_{\mathbf{\hat{e}}_{min}}(x)$ calculates variance along $\mathbf{\hat{e}}_{min}$.
Intuitively, we check that the center of mass is approximately aligned and that the foreground pixels of $o$ lie close to the major axis of $g$.
We empirically find $\epsilon^{\text{SM}}_1 = 10$ pixels and $\epsilon^{\text{SM}}_2 = 1.2$ to yield satisfactory results.

The hardware platform used for inference is a desktop computer with a 4 GHz, 4-core Intel Xeon W-2125 CPU, 64 GB of system memory, and an 11 GB Nvidia GeForce GTX 1080 TI graphics card. The images are captured with an Intel Realsense D455 RGB-D sensor, and low-level control of the Franka Emika Panda robot arm is performed by a separate, realtime-enabled computer with a 3.6 GHz Intel i7-7700 4-core CPU and 16 GB of system memory.

%% file: app/1b_cutting_motion.tex
\subsection{Cutting Motion}
\label{sec:cutting-motion}

\begin{figure}[h]
    \centering
    \includegraphics[width=1.0\textwidth]{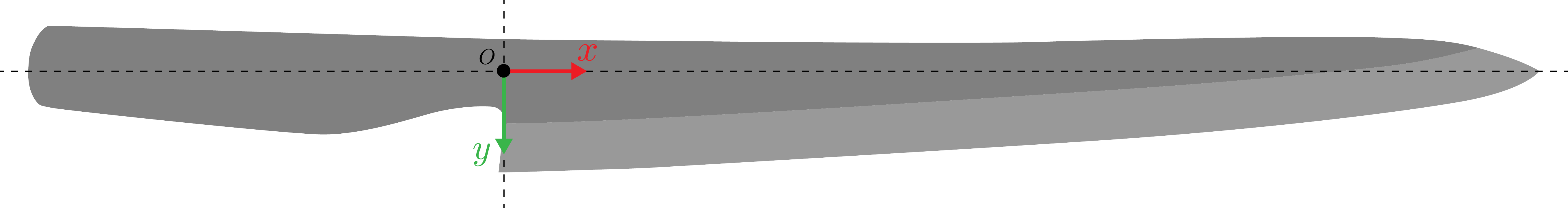}
    \caption{Sashimi knife model used in experiments. ``o'' is the origin of the knife frame (z-axis points into the diagram).}
    \label{fig:knife_model}
\end{figure}

\begin{algorithm}
\caption{Cutting Motion Planning}\label{alg:cutting_motion}
\renewcommand{\algorithmicrequire}{\textbf{Input:}}
\renewcommand{\algorithmicensure}{\textbf{Output:}}
\begin{algorithmic}[1]
\Require initial cutting contact pose w.r.t. the world frame $\mathbf{T}^{init}_{world} \in SE(3)$; cutting profile $\theta[]$ (defining the knife with its model and its blade region to use, and specifying each cutting segment with direction, length, depth, and tilted angle).
\Ensure trajectories for the cut $\xi[]$
\State $L \gets$ the length of $\theta[]$
\State $i \gets 0$
\State $\mathbf{T}^{init'}_{world} \gets \mathbf{T}^{init}_{world}$ \Comment{Initialize current cutting contact pose}
\For{\texttt{$i < L$}}
    \State \texttt{$\xi[i]$, $\mathbf{T}^{init'}_{world} \gets$ PlanTangentPath($\mathbf{T}^{init'}_{world}$, $\theta[i]$)} \Comment{Plan each component path}
    \State \texttt{$i \gets i+1$}
\EndFor
\end{algorithmic}
\end{algorithm}
The cutting motion can be configured with rotation of the knife around its x-axis (tilt angle), see Fig.~\ref{fig:knife_model}, and composed by multiple trajectories $\xi[]$ (e.g. backward-only or forward-and-backward), whose planning is shown in Algorithm~\ref{alg:cutting_motion}. With the input of initial cutting contact pose $\mathbf{T}^{init}_{world}$ and cutting profile $\theta[]$ (specifying each cutting segment with direction, length, depth, tilted angle, etc), each trajectory $\xi[i], i \in [1, \ldots, L]$ is planned as defined by each segment profile $\theta[i]$ via \texttt{PlanTangentPath}($\cdot$). Each component trajectory is planned to ensure the knife blade is tangent to the cutting board at the contact point.
For each point in a trajectory $\xi[i]$, the planned pose of the knife frame w.r.t. the world frame $\mathbf{T}^{knife}_{world}$ is defined as:
\begin{equation}
    \mathbf{T}^{knife}_{world}= (\mathbf{T}^{curr}_{knife})^{-1} \mathbf{T}^{curr}_{init'} \mathbf{T}^{init'}_{world},
\end{equation}
where $\mathbf{T}^{curr}_{init'}$ defines the current contact pose w.r.t. the initial contact pose for the current segment $\mathbf{T}^{init'}_{world}$, which takes into account the tilted angle and the sliding (cut direction and length) and downwards (cut depth) motions of the knife w.r.t. the cutting board; $\mathbf{T}^{curr}_{knife}$ is the current contact pose w.r.t. the knife frame which considers the contact point changes (along the knife blade region to be used for the cut) on the knife and makes sure the knife blade is tangent to the cutting board regardless of tilted angles. $\mathbf{T}^{init}_{world}$ is obtained via the visual perception system (Fig.~\ref{fig:pipeline}a), which is normally the top edge of an object to be cut, with its x-axis indicating the cut direction. The initial cutting contact pose $\mathbf{T}^{init'}_{world}$ for the first component trajectory is $\mathbf{T}^{init}_{world}$, and will be updated accordingly together with the \texttt{PlanTangentPath($\cdot$)} for the rest of the segments.

The Sashimi knife used in the experiments is shown in Fig.~\ref{fig:knife_model}, where the origin of the knife frame is ``o'' with its x-axis pointing from its handle to tip and the y-axis downward. When being used in the experiments, the slightly curved blade of the knife is interpolated from a sparse set of points. 

While Robot A executes the cutting motion, Robot B stabilizes the loin by gently pushing down on it with a fixed pose.
The pose is trivially calculated with a fixed offset from the initial contact pose $\mathbf{T}^{init}_{world}$, which is already aligned with pose of the straightened loin. 
We add a small translational offset to create a safety margin between the holding robot's end-effector and the cutting plane, resulting in about 2-3 cm of the loin extending beyond its grasp. 
Depending on the configured thickness of the slice, the holding robot's pose is periodically moved away from the cutting plane to maintain the desired safety margin.

The cutting planner is executed on the same hardware platform as used for the shape manipulation system (Appendix \ref{sec:method-rl-based-manipulation}). An additional realtime-enabled computer is used for low-level control of the stabilizing arm. It has a 2.2 GHz, 6-core Intel i7-8750H CPU and 32 GB of system memory.

%% file: app/1c_gelsight_feedback.tex
\subsection{GelSight Feedback}
\label{sec:gelsight-feedback-appendix}

We use supervised learning to train a deep neural network to detect whether the knife is in contact with the cutting board based on readings from GelSight sensors.
GelSight sensors are advanced optical tactile sensors with a soft, compliant elastomer surface and embedded with optical structures. 
The elastomer gel pads are coated with reflective materials like silicone rubber or dark pigment. 
When in contact with an object, these elastomer surfaces deform, causing localized changes in reflection intensity. 
A high-resolution camera positioned above the pad captures this deformation pattern.
In our experimental setup, the sensor is mounted on the QB hand holding a knife such that the gel pad of the sensor is in direct contact with the blunt top of the knife blade (see Section \ref{sec:system}).
We also experimented with an additional sensor placed towards the rear end of the handle, but found its data redundant and inferior to those of the front-mounted one.
The sensor maintains continuous contact with the blunt top-edge of the knife throughout the entire cutting motion, enabling recording of gel pad deformations throughout the entire cutting sequence, which forms the basis of our training dataset.

We record and annotate 157 cutting motions, corresponding to a total of 12,397 sensor readings. 
The annotation process is semi-automated with an electronic contact sensor.
Using an Arduino Micro, we connect 3.3V to the knife blade, and connect ground to a sheet of aluminum foil placed on the cutting board with a 10 $k\Omega$ resistor in between.
This turns the knife and cutting board into an electronic switch, allowing us to accurately pinpoint the moment of contact by monitoring voltage over the resistor. 
The salmon loin is somewhat conductive, resulting in a distinguishable, half-high voltage reading of around 1.5V when the blade touches the salmon but not the board.
Although we are mainly interested in contact with the board in this study, the same sensor setup can be used to annotate data for a three-way classification (air vs salmon vs board).
We automatically find the timestamps of contact with salmon and contact with the cutting board by differentiating the voltage time series, finding the two largest peaks with at least 0.5 second separation, and sorting the result (contact with salmon always precedes contact with the board).
This method yields correct results in the vast majority of  cases, but the output is further verified and corrected by a human annotator who inspects voltage readings juxtaposed to footage from a web camera.
Of the 157 annotated cutting trajectories, 137 are used for training and model selection, while the remaining 20 are held aside for testing (used for reported classification metrics in Section \ref{sec:results-tactile}).
In addition, we collect another 20 testing trajectories with the final model in the loop and the electronic contact sensor removed (used for the results seen in Figure \ref{fig:gelsight-results}b).
	
\begin{figure}[ht]
    \centering
    \includegraphics[width=1.\textwidth]{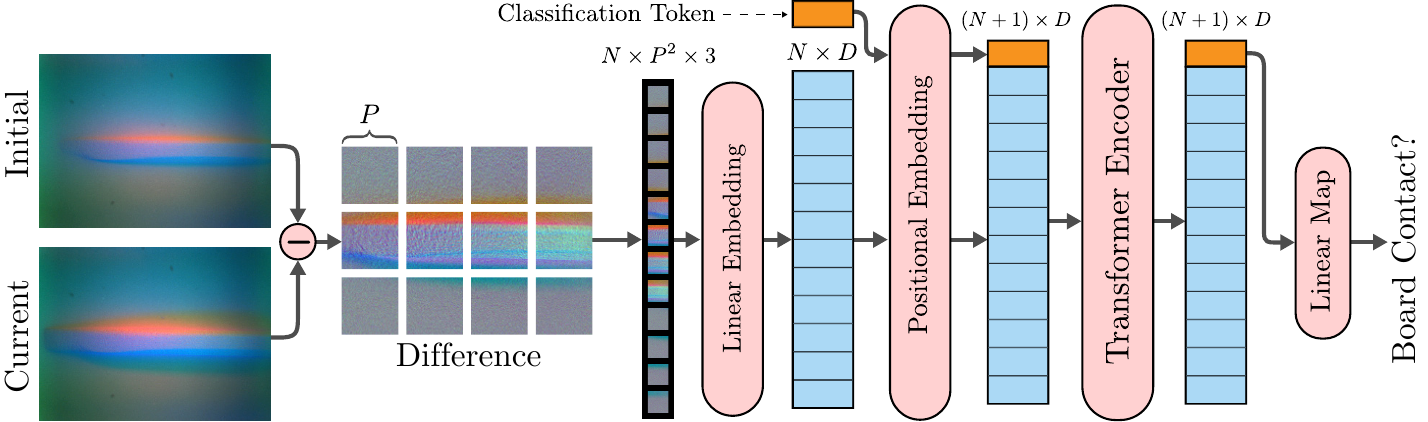}
    \caption{Architecture of the Vision Transformer used to classify gelsight images. Beginning at the left, we take the current GelSight image and subtract one taken at the start of the cutting trajectory (assumed to not be in contact). We then partition the difference image into $N$ patches of $P \times P$ RGB pixels. A linear embedding map is applied to each patch to obtain $N$ patch tokens of dimensionality $D$. A special classification token with learnable values is prepended to the patch tokens. Learnable positional embeddings are added to each token before mixing them in a Transformer Encoder network. Finally, the classification token is extracted and processed with a Linear map to parametrize a (binary) softmax distribution.}   
    \label{fig:gelsight-architecture}
\end{figure}

We train a very lightweight neural network based on the Vision Transformer (ViT) architecture \cite{dosovitskiy2020image} from scratch.
Several models were evaluated in preliminary experiments, including pretrained RESNET50 \cite{he2015deep}, MobileNetV2 \cite{sandler2019mobilenetv2}, and ConvNext \cite{liu2022convnet}.
Despite comparable validation performance, the pretrained models required significantly longer training time and exhibited higher latency for online predictions, which is why we ended up training a much smaller model from scratch.
Our final model can be seen in Figure \ref{fig:gelsight-architecture}. 
The raw GelSight Images are resized from $\sim$4k resolution to aspect-preserving 240 $\times$ 320 pixels before being center cropped to 160 $\times$ 240 (random crop during training).
We found it advantageous to subtract a neutral baseline image, captured at the start of the cutting trajectory, from the current sensor reading. 
The difference is partitioned into $N = 24$ patch tokens of size $P \times P$ = $40 \times 40$ pixels that are linearly mapped to $D=64$ dimensional vectors.
The reminder of the model is a 6-layer ViT classification network with learnable positional embeddings, 6 attention heads, and a feedforward dimensionality of 4.

During training, we augment the image inputs with matching randomized rotations (uniformly $\pm 10^{\circ}$) and translation (uniformly random crops instead of center crops), and use a dropout rate of 0.01 in the ViT.
The network is optimized by minimizing categorical cross-entropy with a Schedule-Free variation \cite{defazio2024road} of the AdamW optimizer \cite{loshchilov2017decoupled}.
We use a learning rate of $3\times10^{-4}$ and weight decay coefficient of 10.0, a batch size of 128, and train for a single epoch.
The reason why we only train for one epoch is that there is high autocorrelation within each cutting trajectory (sampled at $\sim$18Hz), meaning that each datapoint is highly similar to its temporal neighbors.
The network is trained on a desktop computer with a 3.5 GHz 8-core Intel i9-11900k processor, 64 GB of system memory, and a 24 GB NVIDIA GeForce RTX 3090 GPU. Each training run takes only a couple of minutes. We deploy the model for inference on a 2019 MacBook Pro laptop with a 2.6 GHz 6-core Intel i7 processor and 16 GB of system memory. The full sensor loop runs at $\sim$19 Hz. However, most of the time can be attributed to capturing and reading raw 4k images from the GelSight sensor -- the transformer model itself only takes $\sim$4 ms (250 Hz) to evaluate on the modest MacBook platform without any hardware acceleration.

When premature contact with the board is detected with the model, we can adjust the remaining cutting trajectory (Section \ref{sec:cutting-motion}) to prevent further penetration. Suppose a contact is detected at time $t$ and the current knife pose is $\mathbf{T}^{knife}_{world}[t]$. The first thing we need to do is infer the ground plane from the current knife pose. We describe the ground plane with an origin $\mathbf{b} \in \mathbb{R}^3$ and a unit normal vector $\mathbf{\hat{n}_b}\in \mathbb{R}^3$. We assume that the normal vector is approximately known (positive z in our case with a level table), but the exact origin is uncertain. Since we know that the knife just entered contact, the point on the knife that is closest to the ground plane should correspond to a point in the ground plane, meaning that we can use it as an origin. Using our known model of the knife (Fig. \ref{fig:knife_model}), the points along the edge of the knife are described by points $\mathbf{k}^{knife}[j] \in \mathbb{R}^3$ and surface normals $\mathbf{\hat{n}_k}^{knife}[j] \in \mathbb{R}^3$, with $j$ indicating the index of the point. Because the knife is in motion, the world coordinates of the model points vary with time. Let $\mathbf{k}[j, t] = \mathbf{k}^{world}[j, t] = \mathbf{T}^{knife}_{world}[t] \mathbf{k}^{knife}[j]$ and similarly $\mathbf{\hat{n}_k}[j, t] = \mathbf{\hat{n}_k}^{world}[j, t] = \mathbf{R}^{knife}_{world}[t] \mathbf{k}^{knife}[j]$, where $\mathbf{R}^{knife}_{world}[t]$ is the rotation component of $\mathbf{T}^{knife}_{world}[t]$. We can now set the origin $\mathbf{b}$ equal to the knife point $\mathbf{k}[j', t]$, where $j'$ is given by:

\begin{equation}
	\mathbf{b} = \mathbf{k}[j',t] \quad j' = \arg\min_j \mathbf{\hat{n}_b}^T\mathbf{k}[j, t]
\end{equation}

Once we have an origin for the cutting board, we can adjust all future knife poses $\mathbf{T}^{knife}_{world}[t']$, $t' > t$, so that no point of the knife will ever penetrate the cutting board plane. For each future time step $t'$, we find the index $j'$ that correspond to the knife point with the most penetration in the same way as we estimated the cutting board plane origin. We then check if this point penetrates the plane by testing if $\mathbf{\hat{n}_b}^T(\mathbf{k}[j', t'] - \mathbf{b}) < 0$. If so, we generate an adjustment. The adjustment is generated as a translation in the direction of the knife normal $\mathbf{\hat{n}_k}[j',t']$, since we do not want to generate any movement perpendicular to the current cutting direction.  In other words, we want to generate a translation $a \mathbf{\hat{n}_k}[j',t']$, where $a$ is a scalar scaling coefficient so that:

\begin{equation}
\mathbf{\hat{n}_b}^T(\mathbf{k}[j', t'] + a\mathbf{\hat{n}_k}[j',t'] - \mathbf{b}) = 0
\end{equation}

This is straightforward to compute and resolves to:

\begin{equation}
a = \frac{\mathbf{\hat{n}_b}^T(\mathbf{b} - \mathbf{k}[j', t'])}{\mathbf{\hat{n}_b}^T\mathbf{\hat{n}_k}[j',t']}
\end{equation}

Finally, we update the planned trajectory with the adjusted pose $\mathbf{T}^{knife'}_{world}[t']$ by forming a corrective translation transform from $a \mathbf{\hat{n}_k}[j',t']$.

\begin{equation}
\mathbf{T}^{knife'}_{world}[t] = translate(a \mathbf{\hat{n}_k}[j',t']) \mathbf{T}^{knife}_{world}[t]
\end{equation}

%% file: app/1d_vs_based_grasping.tex
\subsection{VS-Based Grasping}
\label{sec:vs-based-grasping}

The controller governing the movement of the sashimi-picking Robot C is based on visual servoing. 
It is initiated with a pickup command sent from the knife-wielding Robot A immediately following a cut.
The initiation command includes a search region, parametrized as a box in 3D space (3-DoF extents + 6-DoF pose). 
The end effector is moved to the top of the box with cartesian position control, and an initial image of the scene is captured by the wrist-mounted RGB-D camera. 

From the initial image, we attempt to locate the newly cut sashimi slice. 
We begin by applying a band-pass filter in HSV space to get a semantic segmentation mask.
The binary mask is further simplified with a morphological denoising filter.
This leaves two large connected foreground objects; the newly cut slice and the remaining part of the main loin. 
Simply picking the second largest object works most of the time but occasionally fails when the main loin gets too small or occluded.
Therefore, we further mask the segmentation image by only considering the pixels corresponding to 3D points inside the search region. 
The search area always excludes the main loin since it is explicitly generated on the opposite side of the cutting plane. 
Finally, we are left with a segmentation image that should only contain the sashimi slice. 
Occasionally, the slice gets stuck to the knife, making it not appear in the initial search area. 
If so, this information is reported back to Robot A, which will then bring the knife into the workspace of Robot C, along with a new search box placed on top of the knife's blade. 
When the slice is found, we commence closed-loop visual servoing.

\newcommand{\mFeat}{\ensuremath{{\mathbf{s}}}}
\newcommand{\mFeatRef}{\ensuremath{{\mathbf{s^*}}}}
\newcommand{\mErr}{\ensuremath{{\mathbf{e_s}}}}
\newcommand{\mVel}{\ensuremath{{\mathbf{v}_e}}}
\newcommand{\mJac}{\ensuremath{{\mathbf{J_s}}}}
\newcommand{\mMatL}{\ensuremath{{\mathbf{L_s}}}}
\newcommand{\mMatV}{\ensuremath{{^c\mathbf{V}_e}}}
\newcommand{\mMatP}{\ensuremath{\mathbf{P}}}
\newcommand{\mMatR}{\ensuremath{{{^c\mathbf{R}_e}}}}
\newcommand{\mVect}{\ensuremath{{^c\mathbf{t}_e}}}

The visual servoing controller moves the end-effector with the aim of grasping the sashimi slice at its center of mass around its major axis. 
For our 4-DoF actuation scheme, we use visual features $\mFeat = (x, y, z, \alpha)$ that allow us to compute a servo law for 3D translation and rotation around the down direction (parallel to the image plane normal vector).
We first generate a feature based on the centroid of the segmentation mask's foreground pixels. 
This provides a translation target in the camera's local xy-plane. 
For the z-component, a common solution is to use the distance from the RGB-D image's depth channel around the centroid.
However, we instead us forward kinematics to calculate the distance from the camera origin to the bottom plane of the search box along the camera forward axis.
This approach has two advantages: First, the forward kinematics of our robot is more precise than the depth channel of the camera, giving us less noisy distance estimates. Second, it effectively puts the z-component of our servo target flush with the underlying surface, meaning that we will grasp as deep as possible, regardless of the thickness of the object that is being grasped.
The distance estimate is used as $z$ value for the feature and used in conjuction with camera intrinsics to deproject the centroid coordinates from the image plane to feature values $x$ and $y$ in the camera's local coordinate frame.
For the angular features $\alpha$, we look at the principal axis of the segmentation mask and calculate its angle with the image plane's vertical axis. 
To minimize the amount of rotation needed to satisfy the goal, the raw angle is projected into the $[-\frac{\pi}{2}, \frac{\pi}{2})$ range by adding an appropriate multiple of $\pi$.

The visual servo control law generates velocities $\mVel = (v_x, v_y, v_z, w_z)$ that drive down the error $\mErr = (\mFeat - \mFeatRef)$ and is based on \cite{chaumette2006Visual}.
Here, $\mFeatRef = (x^*, y^*, z^*, \alpha^*)$ are reference values for the visual features where $(x^*, y^*, z^*)$ corresponds to the center point between the chopsticks in the camera frame and $\alpha = 0$ since we want the object's principal axis to coincide with the vertical axis of the image. 
The end-effector velocity~$\mVel$ is related to change in visual feature error 
$\dot{\mErr}$ 
through the jacobian \mJac.
\begin{equation}
    \dot{\mErr} = \dot{\mFeat} = \mJac \mVel \quad \text{with} \quad \mJac = \mMatL \mMatV \mMatP
    \label{eq:vs-jacobian}
\end{equation}

$\mJac$ is composed from the product of the three matrices \mMatL, \mMatV, and \mMatP.
$\mMatL$ is a $4 \times 6$ interaction matrix that relates a velocity screw in the camera frame $\mathbf{v_c}$ to a change in image features.
$\mMatV$ is the adjoint matrix that transforms a velocity screw from the end-effector frame to the camera frame. 
It is computed from the static transform between the end effector and the camera parametrized by the rotation matrix $\mMatR$ and translation vector $\mVect$.
In the expression below $\left[ \mathbf{u} \right]_\times$ denotes the skew-symmetric matrix equivalent of left cross-multiplication with vector $\mathbf{u}$.
Finally, $\mMatP$ maps velocities from our 4-DoF actuation manifold into their equivalent in 6-DoF screw space.
\begin{equation*}
    \begin{split}
    \mMatL &= \begin{bmatrix}
        -1 & 0 & 0 & 0 & -z & y \\
        0 & -1 & 0 & z & 0 & -x \\
        0 & 0 & -1 & -y & x & 0 \\
        0 & 0 & 0 & 0 & 0 & -1
    \end{bmatrix} \mMatV = \begin{bmatrix}
        \mMatR \; & \left[ \mVect \right]_\times \mMatR \\
        0_3    \;  & \mMatR
    \end{bmatrix} \quad \mMatP = \begin{bmatrix}
        1 & 0 & 0 & 0 \\
        0 & 1 & 0 & 0 \\
        0 & 0 & 1 & 0 \\
        0 & 0 & 0 & 0 \\
        0 & 0 & 0 & 0 \\
        0 & 0 & 0 & 1
    \end{bmatrix}
    \end{split}
\end{equation*}

The final expression for calculating $v_e$ is obtained by inverting $\mJac$ 
and multiplying by gain $\lambda$, yielding the following control law:
\begin{equation}
    \mVel = -\lambda \mJac^{-1} \mErr 
\end{equation}

We run the visual servo controller in two stages. In the first stage, we bring the gripper to a pose just above the object to ensure that position in the xy-plane and the orientation $\alpha$ is sufficiently good. In the second, the gripper is brought all the way down and the chopsticks are closed. In both cases, we run the controller in closed loop until the termination criterion  $\vert (e_x, e_y, e_z) \vert_2 \le \epsilon^{\text{VS}}_1 \wedge \vert e_\alpha \vert < \epsilon^{\text{VS}}_2$ is met, with values $\epsilon^{\text{VS}}_1$ = 0.5 mm and $\epsilon^{\text{VS}}_2$ = 1 degree.

After grasping the object, we switch to open-loop control to place the piece on the serving tray. This involves retreating a few centimeters along the end-effector's z-axis, moving to an intermediary keyframe in joint space, calculating the drop-off pose by indexing a virtual grid placed over the serving tray, and moving there with cartesian position control to deliver the payload. Once the piece has been dropped off, the robot returns to its home keyframe in joint space and awaits the next pickup command.

The controller is executed on a desktop computer with a 2.2 GHz Intel Xeon ES-2630 CPU with 20 cores and 32 GB of system memory. The wrist-mounted camera is an Intel Realsense D405 stereoscopic RGB-D sensor. No hardware acceleration (GPU) is used -- all processing, including image manipulation, happens on the CPU, and the visual servo law runs in closed loop at $\sim$7 Hz.